\pdfoutput=1

\documentclass[11pt]{article}   

\usepackage[]{EMNLP2023}

\usepackage{times}
\usepackage{latexsym}
\usepackage{amsmath}
\usepackage{subcaption}

\usepackage[T1]{fontenc}

\usepackage[utf8]{inputenc}

\usepackage{microtype}

\usepackage{inconsolata}

\definecolor{darkblue}{rgb}{0, 0, 0.5}
\hypersetup{colorlinks=true, citecolor=darkblue, linkcolor=darkblue, urlcolor=darkblue}
\usepackage{booktabs} 
\usepackage{makecell} 
\usepackage{amssymb} 

\usepackage{graphicx}
\usepackage{tabularx, array, multirow}

\title{\textsc{Self-Explore}: Enhancing Mathematical Reasoning in Language Models with Fine-grained Rewards}

\author{Hyeonbin Hwang$^{1}$\hspace{0.7em}Doyoung Kim$^{1}$\hspace{0.7em}Seungone Kim$^{1,2}$\hspace{0.7em}Seonghyeon Ye$^{1}$\hspace{0.7em}Minjoon Seo$^{1}$\\ \\
KAIST AI$^{1}$\hspace{0.5em}Carnegie Mellon University$^{2}$\\
\texttt{\{hbin0701, doyoungkim, seonghyeon.ye, minjoon\}@kaist.ac.kr}\hspace{0.5em}\texttt{seungone@cmu.edu}
}

\begin{document}
\maketitle
\begin{abstract}

Training on large amounts of rationales (\textit{i.e.}, CoT Fine-tuning) has been found effective for improving mathematical reasoning of large language models (LLMs). However, acquiring human-authored solutions or augmenting rationales from proprietary models is costly and not scalable. In this paper, we study the problem of whether LLMs could \textit{self-improve} mathematical reasoning capabilities. To this end, we propose \textsc{Self-Explore}, where the LLM is tasked to explore the first wrong step (\textit{i.e.}, the first pit) within the rationale and use such signals as fine-grained rewards for further improvement. On the GSM8K and MATH test set, \textsc{Self-Explore} achieves 11.57\% and 2.89\% improvement on average across three LLMs compared to supervised fine-tuning (SFT). Our code is available \href{https://anonymous.4open.science/r/Self_Explore-220B}{here}.

\end{abstract}


\section{Introduction}
Recent works have shown that large language models (LLMs) can solve complex reasoning tasks with Chain-of-Thought (CoT) Prompting, which involves generating a rationale before its final prediction~\citep{wei2023chainofthought,kojima2023large,openai2023gpt4, geminiteam2023gemini}. Such ability is especially evident for mathematical reasoning, where many times precise reasoning over multiple steps is required to reach the correct answer ~\citep{fu2023specializing, chen2023program}. Meanwhile, relatively smaller models have shown limited performance, and thus prior works have focused on augmenting rationales from proprietary LLMs and distilling them to smaller models~\citep{li2022explanations, kim2023cot, mukherjee2023orca, yu2023metamath, liu2023tinygsm, mitra2023orca, mitra2024orca, li2024common}. 

However, acquiring high-quality solutions remains challenging. For humans, hand-crafting detailed step-by-step rationale annotations is time-consuming and costly~\citep{kim2023cotever}. On the other hand, using close-sourced models through APIs incurs high expenses and distillation-based methods are inherently limited by the performance of their teacher model, which acts as the upper bound~\citep{gudibande2023false,ye2023flask}. Hence, such strategies are limited in advancing frontier models~\citep{stanton2021does, gudibande2023false}. One potential solution to address this issue is to improve general capability of LLMs through self-training~\citep{gulcehre2023reinforced, chen2024selfplay, yuan2024selfrewarding}.

\begin{figure*}[t]
\centering
\includegraphics[width=1\textwidth,height=\textheight,keepaspectratio]{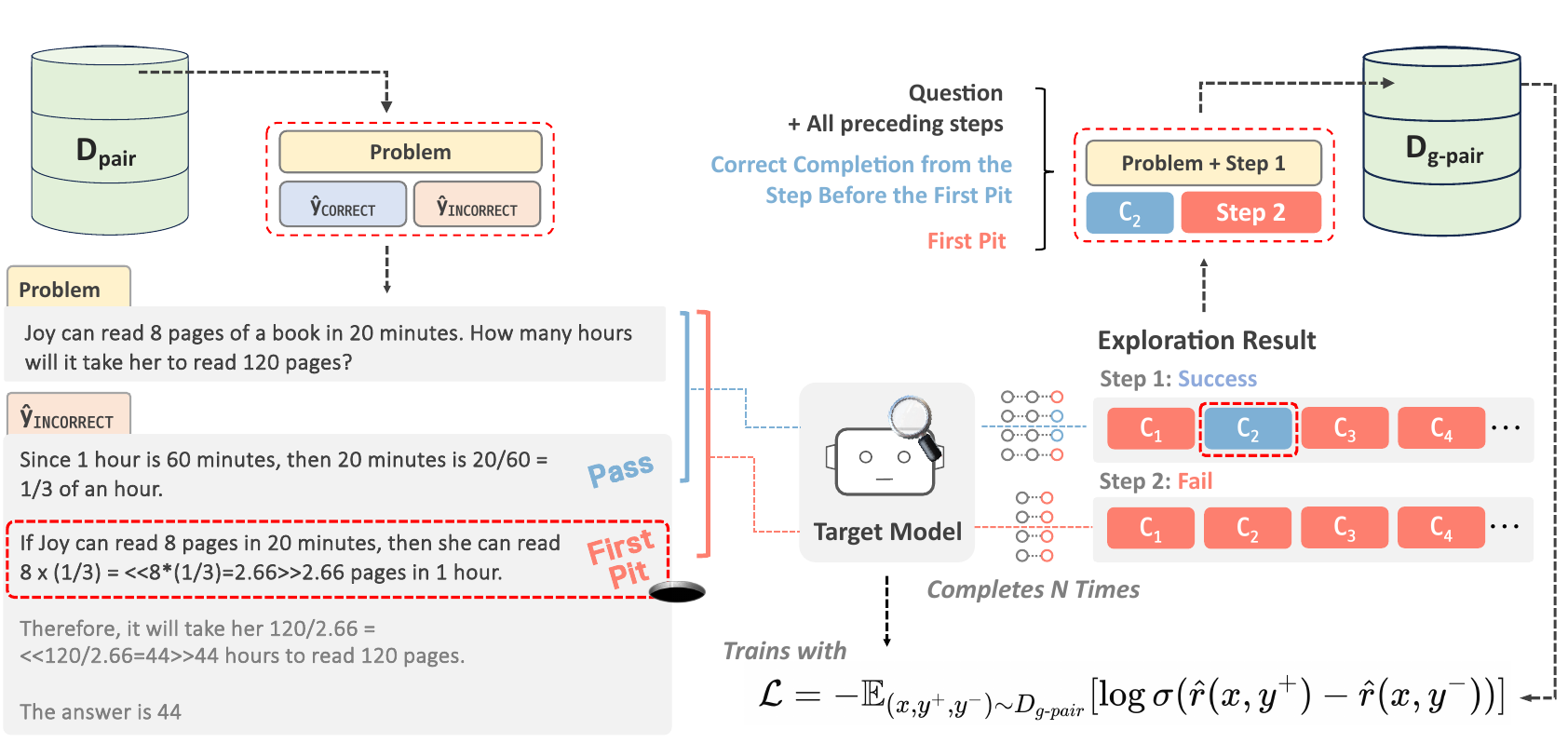}
 \caption{Overview of \textsc{Self-Explore}. From a pairwise dataset ($\mathcal{D}_\mathrm{pair}$) made through outcome supervision, we use the incorrect rationales and make the target model generate multiple completions starting from each step. If none of the completions reach the answer, we mark that step as the \textit{first pit}. Then, with the identified first pit, we reorganize $\mathcal{D}_\mathrm{pair}$ into a granular preference dataset ($\mathcal{D}_{\mathrm{g}\text -\mathrm{pair}}$) which provides better learning signal during training.}
\label{fig:main_figure}
\end{figure*}

Inspired by prior works that focus on aligning LLMs to user preferences through self-training, we propose \textbf{\textsc{Self-Explore}}, a training method designed to self-improve the \textbf{mathematical reasoning} capabilities of LLMs by extracting granular learning signals from its own generated rationales. Specifically, a target model conducts step-level exploration to identify the first wrong step (\textit{i.e.}, first pit) within each rationale by sampling multiple continuations. Then, we construct a pair-wise dataset by sorting the rationales into positive and negative samples at a step level. Finally, by applying an arbitrary preference learning objective (\textit{e.g.}, Direct Preference Optimization (DPO)~\citep{rafailov2023direct}) on a step-level, we relatively increase the probability of generating positive rationales and lower the probability of generating negative ones in a fine-grained manner.

Through experiments, we find that \textsc{Self-Explore} constantly improves the performance across different base models (Mistral-7B~\citep{jiang2023mistral}, Llemma-7B~\citep{azerbayev2023llemma}, and Deepseek-Math 7B~\citep{shao2024deepseekmath}) without any distillation from proprietary models. For each model, we observe a 13.19\%, 10.23\%, and 11.30\% improvement on GSM8K~\citep{cobbe2021training} and a 1.98\%, 3.16\%, and 3.54\% improvement on MATH~\citep{hendrycks2021measuring} compared to supervised fine-tuning (SFT). Also, we find that constructing a pair-wise dataset in a granular manner based on a step-by-step basis (\textit{i.e.}, identifying the first pit) outperforms a naive approach of constructing based on the correctness of the final prediction, leading to a 3.64\% and 2.76\% margin on the GSM8K and MATH dataset, respectively.

\section{Related Works}

\subsection{Mathematical Reasoning of LLMs}

To make a stronger math-reasoning model, previous works have either continually pre-trained the~\textit{base} model on large math corpus~\citep{lewkowycz2022solving, azerbayev2023llemma} or used supervised fine-tuning with a large amount of synthetic dataset distilled from the frontier models~\citep{luo2023wizardmath, yu2023metamath, liu2023tinygsm, mitra2024orcamath, shao2024deepseekmath, toshniwal2024openmathinstruct1}. There is also a growing number of works focusing on increasing test-time compute, namely generating multiple rationales then marginalizing over various reasoning paths~\citep{wang2023selfconsistency}, developing either an \textit{outcome-level} or \textit{process-level} separate verifier that could rank the rationales ~\citep{cobbe2021training, lightman2023lets, liu2023tinygsm, hosseini2024vstar} or decoding under the guidance of a value-model~\citep{xie2023selfevaluation, liu2023dont, yu2023outcomesupervised}. Our approach instead focuses on enhancing the model's top-1 performance which reduces test-time computational burden.

\subsection{Step-level Supervision} 
Many studies have suggested the advantages of step-level guidance~\cite{cobbe2021training, lightman2023lets}, yet acquiring such labels is expensive. Thus, concurrent works rely on pseudo labels, evaluating whether the model can reach the correct answer when provided up to each successive step as input~\citep{wang2024mathshepherd, wang2024multistep, jiao2024learning, havrilla2024glore}. However, most of these works leverage acquired labels to train a verifier model, which is either used for PPO \cite{schulman2017proximal} or inference time re-ranking. Our approach does not require any separate module, much simplifying the overall framework.

\subsection{Self-Training for Mathematical Reasoning}

Another line of works focus on self-training methods that compensate for the scarcity of high-quality training data. This includes utilizing self-generated correct rationales for training ~\citep{zelikman2022star, huang2022large, yuan2023scaling, ni2023learning}, and also self-generated incorrect rationales~\citep{havrilla2024glore, hosseini2024vstar} - which can together form a pairwise dataset that can be trained with preference learning techniques, such as Direct Preference Optimization (DPO)~\citep{rafailov2023direct}.

These strategies are particularly effective for many Math Word Problem (MWP) tasks, where models demonstrate a much higher performance when multiple attempts are allowed (pass@k) rather than just one (pass@1)~\citep{havrilla2024teaching}. This indicates that the model indeed has the potential to reach the correct answer, yet its answer distribution is misaligned. Our work aims to more precisely steer this distribution  towards more optimal policy with fine-grained supervision.

\section{Preliminaries}

Given a language model $\pi_\theta$ and a dataset $\mathcal{D}$, \textit{Self-Training} algorithms comprise of two stages: (1) dataset growth, where the dataset $\mathcal{D}$ is augmented with a $\pi_\theta$'s generations, and (2) policy improvement, where the pre-trained model improves human-alignment through preference learning followed by supervised fine-tuning~\citep{gulcehre2023reinforced, yuan2024selfrewarding, chen2024selfplay}. 
Here, we describe two relevant methods that are employed in our framework for self-training.

\subsection{Rejection Sampling Fine-Tuning}
RFT~\citep{yuan2023scaling} is a training method where the pre-trained model $M_\mathrm{PT}$ is allowed to fine-tune on its own correct generations. To do so, we first need a base generator with zero-shot reasoning ability, which is obtained by training $M_\mathrm{PT}$ on the initial dataset $\mathcal{D}$ with the MLE objective: \begin{equation}\mathcal{L}_{\mathrm{MLE}} = -\sum_{i=1}^{|\mathcal{D}|} \log p_{\theta}(y_i | x_i) \label{eq:1}\end{equation} 

With the resulting model $M_\mathrm{SFT}$, we sample N candidate rationales $\hat{y_i}$ for each question with a nonzero temperature $T$ to form $\mathcal{D}_{\mathrm{GEN}} = \{ (x_i, \hat{y}_{i,j})_{j=1}^N \; | \; x_i \in Q \}$. After removing duplicate rationales using heuristics, each solution $\hat{y_{i,j}}$ is labeled as correct or incorrect by extracting  their predicted final answer with extractor function $\mathcal{F}$ and comparing to the actual answer $a_i$. This set of correct rationales forms $\mathcal{D}_{\mathrm{RFT}}$, and $M_{\mathrm{PT}}$ is trained on this dataset with objective in eq.~\ref{eq:1}. We highlight that in domains with a sufficiently large answer space, (\textit{i.e.} numeric), a correct final answer strongly indicates that the rationale is likely error-free, which is a notable advantage of mathematical reasoning tasks.\footnote{In contrast, if the answer space is small (\textit{i.e.} true/false or multiple choice) selecting the correct option does not necessarily guarantee that the rationale is also correct.}

\subsection{Direct Preference Optimization}
DPO~\citep{rafailov2023direct} training requires a pairwise dataset consisting of a chosen completion $y^+$ and a rejected completion $y^-$ for a given input $x$. Its objective relatively increases the log-likelihood of the chosen completion over the rejected one: $$\mathcal{L}_\mathrm{DPO} = -\mathbb{E}[\log \sigma (\hat{r}_\theta(x, y^+) - \hat{r}_\theta(x, y^-))]$$
\begin{equation}\label{eq:2}
\hat{r}_\theta(x, y) = \beta \log \frac{\pi_{\theta}(y\mid x)}{\pi_{\mathrm{ref}}(y\mid x)}
\end{equation}

Here, reference model $\pi_{\mathrm{ref}}$ is generally initialized with supervised fine-tuning (SFT) with preferred completions for a single epoch to minimize distribution shift from the true reference distribution.

\begin{figure}
\centering
\includegraphics[width=\linewidth]{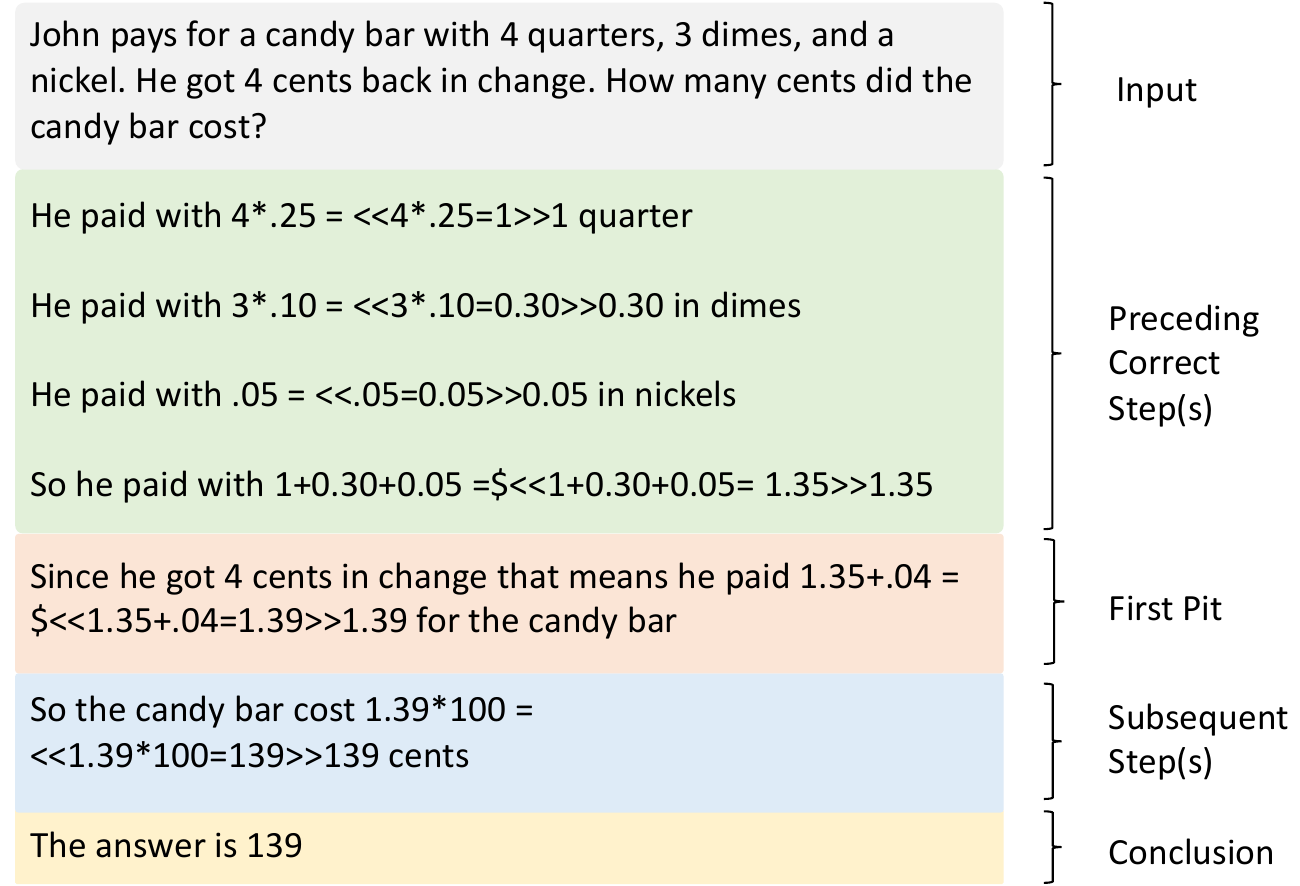}
\caption{Example of a rejected sample from GSM8K: In the First Pit, 0.04 was mistakenly added instead of being subtracted.} 
\label{fig:example_figure}
\end{figure}

\section{Method}

\subsection{Self-Training with Outcome Supervision}
To achieve autonomous improvement for multi-step reasoning, we follow the general offline preference recipe from recent models (SFT + DPO) ~\citep{tunstall2023zephyr, ivison2023camels}. Yet we only utilize the initial human-curated dataset $\mathcal{D}$ and the training model's self-generated data. In this light, we initialize the reference policy $\pi_\mathrm{ref}$ by applying Rejection Sampling Fine-Tuning to the pre-trained model $M_\mathrm{PT}$ to obtain $M_\mathrm{RFT}$.

To construct the pairwise dataset $\mathcal{D}_{\mathrm{pair}}$, we start by adopting the conventional approach of designating a correct solution as a favorable sample ($y^+$) and an incorrect solution as an unfavorable sample ($y^-$) for a given problem $x$, using outcome supervision to determine correctness~\citep{yu2023outcomesupervised, hosseini2024vstar}. We pair each correct solution $\hat{y_{i,j}}$ in $\mathcal{D}_{\mathrm{RFT}}$ with the incorrect solution $\hat{y_{i,k}}$ from $\mathcal{D}_{\mathrm{GEN}}$ that has maximum edit distance in-between, in light of ~\citet{pal2024smaug}. Overall, we utilize each solution only once and continue this pairing process until no further pairs can be formed. For additional details on pair formation, please see Appendix \ref{appendix:pair}. 

After forming the pairwise dataset $\mathcal{D}_\mathrm{pair}$, we train the model $M_\mathrm{RFT}$ using the objective specified in eq.~\ref{eq:2}. This approach guides the model on a \textit{holistic} level to favor policies that generate solutions leading to the correct answer, relative to those that result in an incorrect answer. In the following sections, references to DPO specifically denote this outcome-supervised preference learning approach which we employ as a baseline method for our experiments. 

\subsection{Multi-Step Preference Learning} 
In preference learning, a language model $\pi_\theta$ functions as an agent optimized to generate responses that maximize a given reward function. In a multi-step problem setting, given an input $x$ and a target sequence $y$ comprising steps $\{y^1, ..., y^n\}$, we can define the agent's reward by evaluating the predicted final answer at the terminal state $y^n$ against the ground truth answer $a$ using the extractor function $\mathcal{F}$ .

\begin{equation}
    r(x, (y^1, ... , y^n))=
    \begin{cases}
      1, & \text{if}\ \mathcal{F}(y^n) = a \\
      -1, & \text{if}\ \mathcal{F}(y^n) \neq a \\
    \end{cases}
\end{equation}

If the answer space is large enough, we can safely assume that a match between these two indicates that all the prior steps $\{y^1, ..., y^{n-1}\}$ are also correct. On the other hand, if the terminal state $y_n$ reached an incorrect answer, this suggests that the sequence generation has encountered at least one "\textbf{\textit{pit}}" - an irreversible error in its prior steps that caused the agent to deviate from the correct path. Yet once we identify the \textit{first pit}, we can consider all subsequent steps as non-relevant, given that they are already compromised by the preceding pit. Then, we can re-design the reward for generating each step $y^i$ in the multi-step problem setting as follows:

\begin{equation}
    r(x, (y^1, ... , y^i))=
    \begin{cases}
      -1, & \text{if}\ y^i \text{ is a \textit{first pit}} \\
      1, & \text{if} \; i = n \\
         & \text{and} \; \mathcal{F}(y^i) = a\\
      0, & \text{otherwise} \\
    \end{cases}
    \label{eq:4}
\end{equation}

Meanwhile, the challenge posed by multi-step tasks is that it is hard to avoid the pit and reach the correct terminal state, especially when the problem requires many steps to solve. For simplicity, if we assume there is a constant probability of $\epsilon$ to fall into the pit in each stage, then the expected reward after generating t steps becomes $(1 - \epsilon)^t$, which exponentially decreases as $t$ gets larger. In order to minimize this \textit{risk}, previous works have utilized DPO to enable the original model as a reward model, steering away from the episodes that the model fell into the \textit{pit}. However, DPO objective shown in eq.~\ref{eq:2} relatively decreases the likelihood of all tokens in the rejected solution $y^-$. In light of eq.~\ref{eq:4}, we claim that only the step corresponding to the \textit{first pit} should be discouraged. To elucidate, we consider the following two cases.

\textbf{(1) Steps before the first pit} For a rejected solution $y^- = \{y^1, ..., y^n\}$, there always exists an initial wrong step $y_{\mathrm{w}}$ corresponding to the \textit{first pit}. If $w \neq 1$, the reward of the preceding steps $r(y^i |x, y^1, ... , y^{i-1})$, such that $i \leq w-1$ should not be penalized. 

\textbf{(2) Steps after the first pit} For the steps subsequent to $y_w$, while it's clear that $y^w$ is flawed, decreasing the likelihood of $\sum_{i=w+1}^{n} P(y^{i}| x, y^1, ... , y^{i-1} )$ could adversely impact the coherency of the model. This concern arises because the error in $y^w$ may be due to a minor computation error or wrong formula construction, whereas the subsequent reasoning and steps could still be logically sound. (Figure \ref{fig:example_figure})

\begin{figure*}[t]
\centering
\includegraphics[width=1\textwidth,height=\textheight,keepaspectratio]{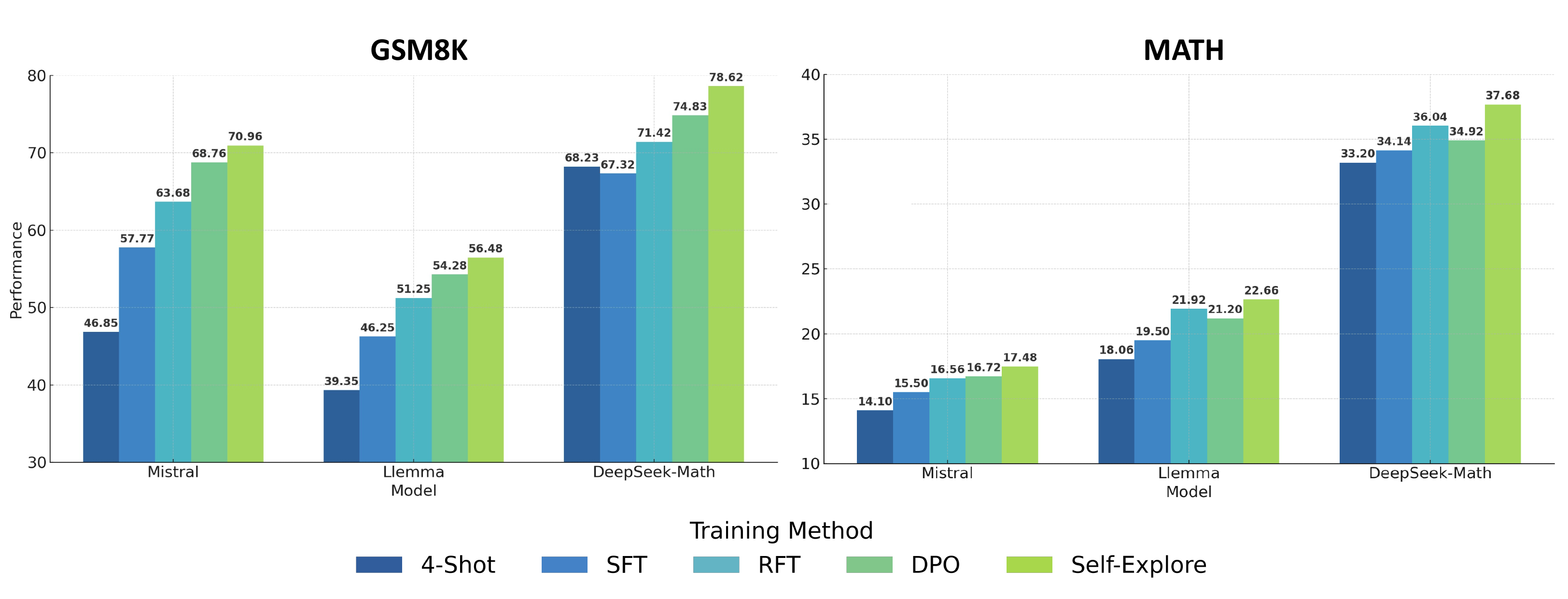}
\caption{Result of three models trained with diverse learning methods. \textsc{Self-Explore} shows consistent superiority over other training methods in both GSM8K and MATH benchmark. For 4-Shot, we report the best performance achieved across three distinct prompts.}
\label{fig:main_result}
\end{figure*}

\subsection{Self-Explore}

In this light, we apply the reward design from eq.~\ref{eq:4} to transform $\mathcal{D}_{\mathrm{pair}}$ into step-level \textbf{granular} pairwise dataset  $\mathcal{D}_{\mathrm{g}\text -\mathrm{pair}}$. This requires modifying each \textit{rejected} sample within $\mathcal{D}_{\mathrm{pair}}$ so that we only reduce the likelihood of the\textbf{ first pit} $y^w$. To find such a step, we employ our model as a self-guided explorer.

We assess whether the target model can reach the correct answer by sampling $k$ completions with a non-zero temperature $T$ from each step. If none of the completions yield the correct answer, we label that step as $y^w$. This indicates that the step has low \textit{Q-value} or potential, suggesting that the step is either incorrect or is beyond the model’s capability to utilize it effectively. On the other, if we do not find $y^w$ until the end, we discard that sample. This is because the absence of $y^w$ suggests that the sample, produced by the base generator ($M_{\mathrm{SFT}}$), may not actually be \textit{infeasible} from the perspective of the explorer ($M_{\mathrm{RFT}}$).

To form a $\mathcal{D}_{\mathrm{g}\text -\mathrm{pair}}$ instance, we set the first pit $s^w$ as the new rejected sample. The new input is then created by concatenating the original input (question) with all steps prior to the first pit. For the new chosen sample, we randomly select one correct completion from the step just before the first pit ($s^{w-1}$), that matches this new input. We intentionally use the \textit{whole} completion from the explorer to maximize the expected learning signal, thus the likelihood of deriving the correct answer. In a similar manner, if $w=1$, we simply use the original chosen sample. Finally, we train with preference learning objective in eq.~\ref{eq:2} on our reference model $M_{\mathrm{RFT}}$ using this new dataset $\mathcal{D}_{\mathrm{g}\text -\mathrm{pair}}$.

Our step-level annotation strategy builds on the framework first introduced in~\citet{wang2024mathshepherd}. However, unlike Wang's approach which utilizes different models for each role (\textit{i.e.} completer, target model, and reward model), our method forms a preference pair using this label which  allows the integration of these distinct systems into a single model, much simplifying the overall training process.

\subsection{Experiments}

\textbf{Datasets and Models} We conduct our experiments on two widely used MWP datasets GSM8K~\citep{cobbe2021training} and MATH~\citep{hendrycks2021measuring}. GSM8K dataset consists of 7,473 training and 1,319 test problems, while the MATH dataset contains 7,500 training and 5,000 test problems. We test \textsc{Self-Explore} across 3 different models: Mistral-7B~\citep{jiang2023mistral}, Llemma-7B~\citep{azerbayev2023llemma}, and Deepseek-Math-7B-Base~\citep{shao2024deepseekmath}.\footnote{All datasets and models are under MIT license, except for Mistral-7B which is under Apache 2.0. We use these solely for research purposes.}

\textbf{Hyperparameters}  For the base generator $M_{\mathrm{SFT}}$, we only train for 2 epochs, yet report the performance of the best checkpoint over 5 training epochs to ensure a fair comparison. Similarly, For $M_{RFT}$, we train the model for one epoch, yet report the best performance achieved over the course of 5 epochs. For all supervised fine-tuning, we use overall batch size of 64 and conduct learning rate search between $\{ 1e^{-6}, 1e^{-5}\}$ for all models. To construct $\mathcal{D}_{\mathrm{RFT}}$, we use $N = 100$, with $T = 0.7$. For step-level exploration, we also use temperature of $0.7$, and generate $k = 4$ at each step. All our generations were carried out using \textit{vllm}~\citep{kwon2023efficient}. For DPO training, we use overall batch size of 32, conduct learning rate search among $\{1e^{-6}, 5e^{-6}, 1e^{-7}\}$, and train for 3 epochs to report the best performance.

\section{Results}
\label{results}

\subsection{Main Results}
As shown in Figure \ref{fig:main_result}, \textsc{Self-Explore} shows the highest performance in MATH and GSM8K compared to other methods. Especially, our method shows 13.19\%, 10.23\%, 11.30\% increase in GSM8K and 1.98\%, 3.16\%, 3.54\% increase in MATH compared to Supervised Fine-Tuning (SFT) for each model. Also, it consistently performs better than training DPO with outcome-supervised rewards from $\mathcal{D}_\mathrm{pair}$, which shows the strength of our step-level reward. 

Meanwhile, DPO performs worse than RFT in MATH dataset for Llemma and DeepSeek-Math. Note that this does not mean that DPO brought performance degradation, but rather RFT (1 epoch) + DPO achieved less performance than the optimal checkpoint achieved by RFT alone. For instance, when DPO was applied to the one-epoch RFT checkpoint, the performance showed a marginal increase from 34.82 to 34.92, whereas applying \textsc{Self-Explore} to the same checkpoint achieved 37.68. Unlike the granular supervision provided by \textsc{Self-Explore}, we hypothesize that outcome supervision offers a significantly weaker training signal. This weaker signal is more challenging for the model to interpret and utilize effectively, making it harder to guide the model towards a successful policy. This may rather lead to reward exploitation or undesired penalization of correct steps that may not necessarily improve its general reasoning ability. 

We also note that the performance gain in MATH is much lower when compared to GSM8K, which is primarily due to its difficulty. Not only the task itself is more inherently challenging, but also the training dataset is limited in size, which is then tested against a large pool of test problems. We hypothesize that low performance of $M_{\mathrm{RFT}}$ as both generator and completer prevents an effective exploration process when conducting both overall generation and step-level search. In fact, for the MATH dataset, we observe number of unique question-level samples in $\mathcal{D}_{\mathrm{RFT}}$ resulting significantly less. For more details about the dataset statistics, please refer to Appendix \ref{appendix:dataset}.

\begin{table}[]
\centering
\small
\begin{tabularx}{\linewidth}{Xcc}
\toprule
{Data Type}               & {GSM8K} & MATH \\
\midrule
Pairwise                      & 74.83  & 34.92  \\
Granular Pairwise         & \textbf{78.47} & \textbf{37.68}    \\
$\;$ - Choose only First Step                 & 75.74 & 35.76   \\
$\;$ - Reject All                & 75.89 & 36.82    \\
\bottomrule
\end{tabularx}
\caption{DeepSeek-Math's GSM8K test set accuracy when trained with DPO on various types of preference data.}
\label{tab:label_design}
\end{table}

\subsection{Step-Level Reward Design} 

To better justify our design for step-level fine-grained reward, we conducted tests on DeepSeek-Math using two additional settings from our current dataset, $\mathcal{D}_{\mathrm{g}\text -\mathrm{pair}}$. 1) Choose Only First Step: For the new chosen sample, we take only the \textit{first correct step}, rather than the \textit{entire} completion. This approach aligns with the new rejected sample, where we only minimize the likelihood of the first \textit{pit} alone. 2) Reject All: For the new rejected sample, we reject the first pit along with its all subsequent steps. We no longer regard the steps after the first pit as irrelevant; instead, we aim to reduce their likelihood as well.

As shown in Table \ref{tab:label_design}, we observe that training with our fine-grained reward yields the best performance in both datasets. While the two other settings perform better than training with outcome-supervised pairwise dataset, they both result in suboptimal performances. This again highlights the idea that the learning signal becomes the most effective when maximally utilizing the whole correct solution while decreasing only the first pit, which is in line with the eq.~\ref{eq:4}.

\begin{table}[htbp]
    \centering
    \small
    \label{tab:method_accuracy}
    \begin{tabularx}{\linewidth}{Xcccc}
        \toprule
        Dataset & $k = 4$    & $k = 8$    & $k = 16$   & $k = 32$   \\
        \midrule
          GSM8K   & \textbf{70.96}& 69.9 & 70.81& 70.05 \\
        MATH    & \textbf{17.48}& 17.4 & 17.44& 17.10 \\
        \bottomrule
    \end{tabularx}
    \caption{Performance of Mistral-7B for GSM8K and MATH datasets, with varying exploration size $k$.}
    \label{tab:exp_size}
\end{table}

\section{Analysis} 

\subsection{Ablation Studies}

\textbf{Effect of Exploration Space} We further analyze whether larger exploration space leads to a better performance. Specifically, we aim to analyze whether steps in the rejected sample which have \textbf{low, non-zero} total expected reward (\textit{i.e.} low probability of reaching to the correct answer) should not be discouraged. These could be found by exploring more paths with larger $k$. On the other, one could argue that it is better to prevent the model from going through such path from the outset by rigorously evaluating each step against a strict standard of smaller $k$. Therefore, we test Mistral-7B with varying step-level exploration size $k$ among $\{4, 8, 16, 32\}$, with which we accordingly build each $\mathcal{D}_{\mathrm{g}\text -\mathrm{pair}}$ and train the target model with the DPO objective.

As shown in Table \ref{tab:exp_size}, we see that increasing exploration size does not lead to performance increase, yet rather often leads to degradation. First pit detection indeed does occur in later stages when using larger exploration space - for instance, for MATH dataset the mean index of $s^w$ becomes $1.86 \rightarrow 2.19 \rightarrow 2.61 \rightarrow 3.13$ with increasing $k$ values. However, this does not necessarily extend to a better resulting model performance.

We believe that while it may be technically \textit{feasible} to reach an answer through a certain step, it does not necessarily mean that it is \textit{favorable}. For instance, if a model has a high probability $\epsilon$ of falling into the pit after a given correct step (\textit{i.e.} it tends to associate post-sequences that is logically incorrect), sometimes it may be more effective to avoid such step from the beginning, if there are other correct alternatives that can lead to the correct answer with less future \textit{risk}. In this manner, we hypothesize it is favorable to optimize the steps with high total expected rewards, or otherwise it may introduce unnecessary noise.

\begin{table}[]
\centering
\small
\begin{tabularx}{\linewidth}{Xc}
\toprule
Method & Acc. \\ 
\midrule
RFT & 63.68 \\  
\addlinespace 
\midrule
DPO & 66.64 \\ 
\addlinespace
\midrule
\textsc{Self-Explore}: Completers \\
Mistral\textsubscript{SFT} & 67.70\\ 
Mistral\textsubscript{RFT} (Ours) & 68.46\\ 
DeepSeek\textsubscript{RFT} & 66.79 \\ 
GPT-4 & \textbf{69.14} \\ 
\bottomrule
\end{tabularx}
\caption{GSM8K Test Set Accuracy of the Mistral-7B when trained DPO with 5.8K instances of  supervised by different completers.}
\label{tab:supervisor}
\end{table}

\textbf{Effect of Explorer} We also investigate the potential of enhancing model performance by adopting a different explorer (or supervisor). Current labeling method guarantees a fairly reasonable step-level accuracy~\citep{wang2024mathshepherd}, yet as $\mathcal{D}_{\mathrm{g}\text -\mathrm{pair}}$ data quality heavily depends on the explorer's capability, we hypothesize that our final model performance may be bottlenecked by the explorer's limitations.   

With this end, we train DPO objective on Mistral-7B $M_\mathrm{RFT}$
with $\mathcal{D}_{\mathrm{g}\text -\mathrm{pair}}$ completed by a range of models, \textit{i.e.} $\textnormal{Mistral}_{\mathrm{SFT}}$, $\textnormal{Mistral}_{\mathrm{RFT}}$, $\textnormal{Deepseek-Math}_{\mathrm{RFT}}$, and GPT-4 \citep{openai2023gpt4}. 
We use the same step-level exploration approach in \textsc{Self-Explore} except for GPT-4, which showed tendency to identify the wrong step instead of completing from the given steps even when provided with explicit instructions. Therefore, we directly prompted GPT-4 to pinpoint the first wrong step and to generate correct sequence from there while ensuring it maintains the original style of the preceding steps. To leverage GPT-4 as the oracle completer, we curated a specialized subset of $\mathcal{D}_{\mathrm{pair}}$ to start with. We first chose one sample per each unique problem $x_i \in \mathcal{D}_{\mathrm{pair}}$, and only included samples where GPT-4 successfully arrived at the correct conclusion, resulting in total of 5.8K samples. 

As shown in Table \ref{tab:supervisor}, we see applying DPO with either $\mathcal{D}_\mathrm{pair}$ and $\mathcal{D}_{\mathrm{g}\text -\mathrm{pair}}$ results in lower performance due to the dataset's smaller size. Yet, we observe that \textsc{Self-Explore} still performs better than outcome-supervised DPO in small-scale. Also, while DeepSeek\textsubscript{RFT} itself performs better as a \textit{generator} than Mistral\textsubscript{RFT} (\textit{i.e.} 71.42 vs 63.68), as a completer for Mistral\textsubscript{RFT}, the former yields higher efficiency. We deduce this may be due to the fact that DPO generally works better when the training data, especially when the chosen completions are closer to its distribution, which is also suggested by the common practice of training SFT for one epoch prior to DPO~\citep{rafailov2023direct, yuan2024selfrewarding}. 

Finally, we observe that using oracle completer GPT-4 results in a better final model performance than using the same model's $M_\textsubscript{RFT}$. We believe that as the generated completions by GPT-4 does not fully represent the target model's distribution, if the completions were generated by a hypothetical \textit{oracle} $M_\textsubscript{RFT}$ of the same model, performance would have been even higher. We believe that this suggests that our method could be further improved with more robust exploration methods.

\textbf{Effect of Objective Function} We also analyze whether the effectiveness of our fine-grained data can be extended to other preference learning objectives, such as IPO~\citep{azar2023general} and KTO~ \citep{ethayarajh2024kto}. With other settings equal, we train Mistral-7B's $M_\mathrm{RFT}$ using $\mathcal{D}_\mathrm{pair}$ and 
$\mathcal{D}_{\mathrm{g}\text -\mathrm{pair}}$, for 1 epoch and $\tau=0.01$ for IPO. 

\begin{figure}
\centering
\includegraphics[width=\linewidth]{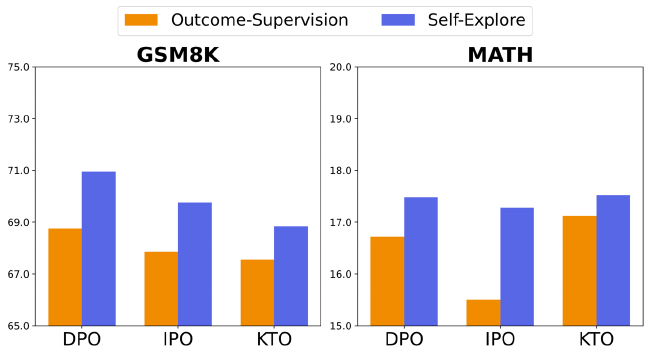}
\caption{Mistral-7B performance when trained  with different preference learning objectives using outcome-level supervision ($\mathcal{D}_\mathrm{pair}$) or \textsc{Self-Explore} ($\mathcal{D}_{\mathrm{g}\text -\mathrm{pair}}$)} 
\label{fig:pl_methods}
\end{figure}

In Figure~\ref{fig:pl_methods}, we see that for both datasets using fine-grained supervision consistently results in better model performance than using outcome-supervised pairwise data. This shows the robustness of \textsc{Self-Explore} across various objectives, highlighting the general effectiveness of our fine-grained data. We have also experimented using high values of $\tau$ for IPO and ORPO~\citep{hong2024orpo}, however they showed degraded performance for both types of supervisions.\footnote{We posit that the efficacy of self-training hinges on the introduction of a strong distinct positive signal for the chosen examples and negative signal for the rejected ones.}

\subsection{Qualitative Analysis}

We also qualitatively analyze whether the numerical performance gains also translate into improved solution quality. To do so, we randomly select 100 questions from GSM8K\footnote{We use GSM8K to guarantee a robust evaluation performance of GPT-4.} Test set and generate response from DeepSeek-Math models trained with RFT, DPO, and \textsc{Self-Explore}. Then, we use GPT-4 as our evaluator using FLASK~\citep{ye2023flask}, effectively assessing the given solution's logical robustness, efficiency and correctness in a scale of 1-5 against the ground truth solution.

As shown in the Table \ref{tab:logical}, we see that \textsc{Self-Explore} scores the best result in all criteria. Also, the general trend in the table implies that increased numerical performance does indicate a better quality in terms of correctness, robustness, and efficiency. We hypothesize that our method guides the model to better utilize its available knowledge, leading to the generation of solutions that are both more efficient and robust. For additional details and examples on FLASK evaluation, please see Appendix \ref{appendix:flask}.

\section{Conclusion}

In this paper, we propose \textsc{Self-Explore} where LLMs can self-improve from given initial human curated dataset using fine-grained supervision. By utilizing automatic self-exploratory annotation, \textsc{Self-Explore} effectively integrates the roles of the annotator, target, and reward models into a single system. On mathematical reasoning datasets GSM8K and MATH, our method outperforms traditional supervised fine-tuning (SFT) method by 11.57\% and 2.89\% in average across three different models, respectively. Furthermore, we demonstrate that our method introduces minimal computation overhead (See Appendix \ref{appendix:compute}). We hope our work could motivate future works to explore self-training methods that could more robustly generalize to a broader reasoning space across various domains, with ends of advancing the frontier of LLM reasoning. 

\renewcommand{\arraystretch}{1.2}
\begin{table}[ht]
\label{tab:4}
\small
\centering
\begin{tabularx}{0.48\textwidth}{@{}Xccc@{}}
\toprule
\textbf{Model}        & \textbf{Robustness} & \textbf{Correctness} & \textbf{Efficiency} \\ \midrule
RFT                   & 3.87                        & 3.86                         & 4.07                        \\
DPO                   & 4.19                        & 4.15                         & 4.35                        \\
Self-Explore         & \textbf{4.27 }                       & \textbf{4.28 }                        & \textbf{4.44 }                       \\ \bottomrule
\end{tabularx}
\caption{Comparison of FLASK Logical Metrics Across Different Training Methods, using DeepSeek-Math on GSM8K.}
\label{tab:logical}
\end{table}

\section*{Limitations}

We propose a method on how to better exploit the \textit{solution} space to provide a better fine-grained supervision for self-improving reasoning capabilities. Yet given limited amount of questions, which is a quite common scenario, preference learning with self-generated samples may be prone to overconfidence and thus increases top-1 performance at the expense of diminished test-time exploration robustness~\citep{cao2024scalable}. We suspect this is related to reward overoptimization~\citep{gao2022scaling,burns2023weaktostrong} and attach relevant analysis in Appendix. We leave as a future work on methods for mitigating this overoptimization, where one promising direction could be exploring the potential of integrating collection of diverse datasets as in~\citet{longpre2023flan}, so that the model can generalize across a broader question space.

Also, our work is currently conducted with 7B pre-trained models and does not consider extensively fine-tuned CoT models or larger scale architectures that have shown stronger reasoning capabilities~\citep{yu2023metamath, mitra2024orcamath, shao2024deepseekmath}. We believe for practical self-training applications, it is crucial to explore continual training processes on these sophisticated models. While this paper aims to compensate for distilled rationales used in instruction-tuned models, we encourage future works to investigate about such models could further benefit from self-improvement processes in a robust and effective manner.

\section*{Acknowledgements}
This work was partly supported by the LG AI Research grant (Self-improving logical reasoning capabilities of LLMs, 2022, 50\%) and the Institute of Information \& Communications Technology Planning \& Evaluation(IITP) grant funded by the Korea government(MSIT) (RS-2024-00397966, Development of a Cybersecurity Specialized RAG-based sLLM Model for Suppressing Gen-AI Malfunctions and Construction of a Publicly Demonstration Platform, 50\%).

\bibliography{anthology,custom}
\bibliographystyle{acl_natbib}

\newpage

\begin{figure*}[t]
\centering

\begin{subfigure}{\textwidth}
    \centering
    \includegraphics[width=0.88\textwidth,height=\textheight,keepaspectratio]{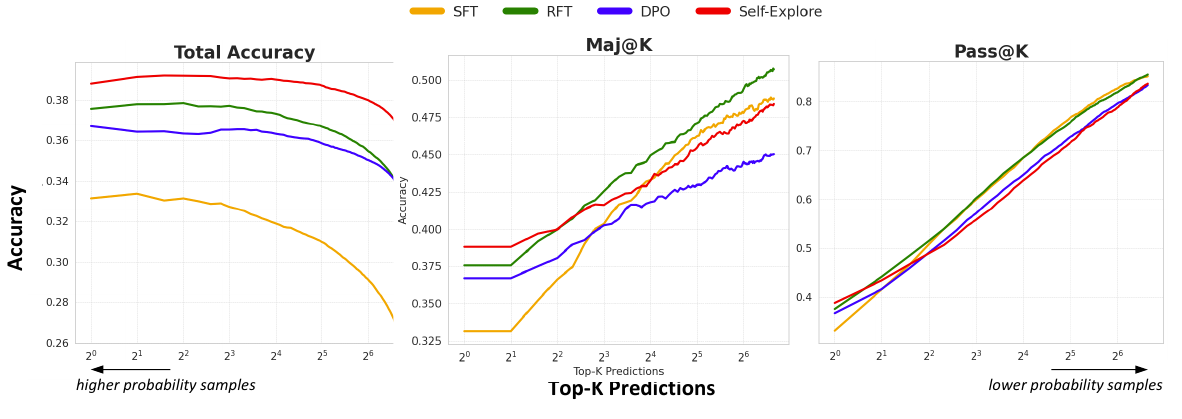}
    \vspace{-3mm}
    \caption{Performance on \textbf{MATH}}
\end{subfigure}
\vspace{2mm}

\begin{subfigure}{\textwidth}
    \centering
    \includegraphics[width=0.88\textwidth,height=\textheight,keepaspectratio]{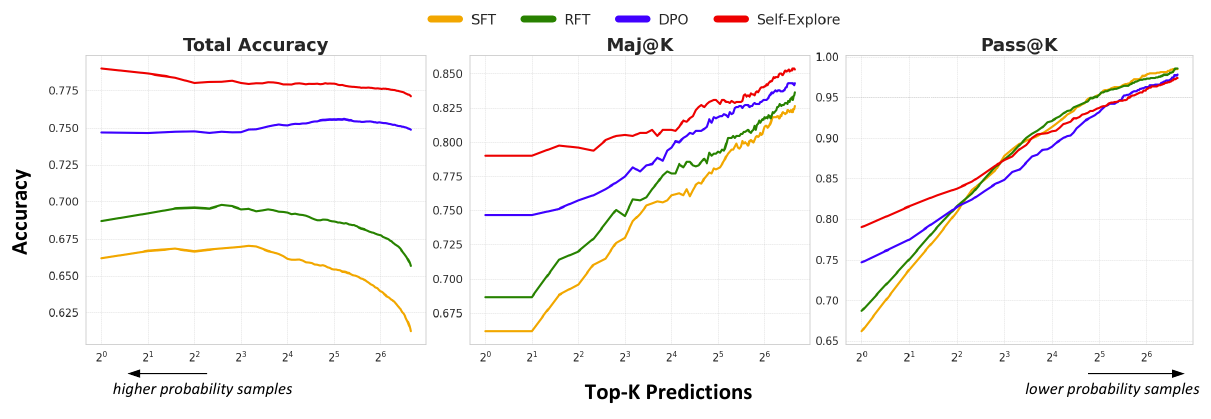}
    \vspace{-1.7mm}
    \caption{Performance on \textbf{GSM8K}}
\end{subfigure}

\caption{Performance of DeepSeek-Math Model on different datasets when trained with diverse training methods - we report using three metrics: total accuracy, maj@k (\textit{i.e.} self-consistency), and pass@k.}
\label{fig:passk}
\end{figure*}

\appendix

\section{Post-Training Distribution}
\label{appendix:post_training}

Here we analyze how the model's distribution changes after applying different training methods, including RFT, DPO and \textsc{Self-Explore}. Specifically, we use our best performing model DeepSeek-Math with a special focus on MATH dataset to explore potential directions on how LLMs could better self-improve in more advanced reasoning capabilities.

While we previously used greedy decoding to report the top-1 performance, here we sample 100 predictions per problem from the test set with temperature of $0.7$ and sort the generations by the overall sequence likelihood in descending order. Then, we report its performance in three different metrics, total accuracy, self-consistency (maj@k), and pass@k, in Figure \ref{fig:passk}.

For the total accuracy, we observe a general trend of curves decreasing with the inclusion of samples with lower overall likelihood. Yet, we observe DPO and \textsc{Self-Explore} displaying smaller gaps of reduction. Numerically, as $k$ goes from $1$ to $100$, \textsc{Self-Explore} performs 0.388 $\rightarrow$ 0.367, DPO 0.367 $\rightarrow$ 0.337, and RFT 0.376 $\rightarrow$ 0.336. We believe preference learning with self-generated samples minimizes the \textit{risk} as even token generation with comparatively lower likelihood to be sampled eventually lead to the correct answer.

However, this comes at the cost of reduction in sample diversity where preference learning (\textit{i.e.} RLHF) has been previously reported to promote similar phenomena~\citep{kirk2024understanding}. We believe this even intensifies as we are training with our own generated data. To support this, we leverage BERT~\citep{devlin2019bert} to extract embeddings of model generations and express solution diversity as the average per-input pairwise distance of embeddings, \textit{i.e.} for $i^{th}$ sample, this is given as: $$d_i =  \frac{2}{N \cdot (N - 1)} \sum_{j=1}^{N-1} \sum_{k=j+1}^{N} d(h_{i,j}, h_{i,k})$$

where $h$ is the embedding and $N = 100$ in this case. 

\begin{figure}
\centering
\includegraphics[width=0.9\linewidth]{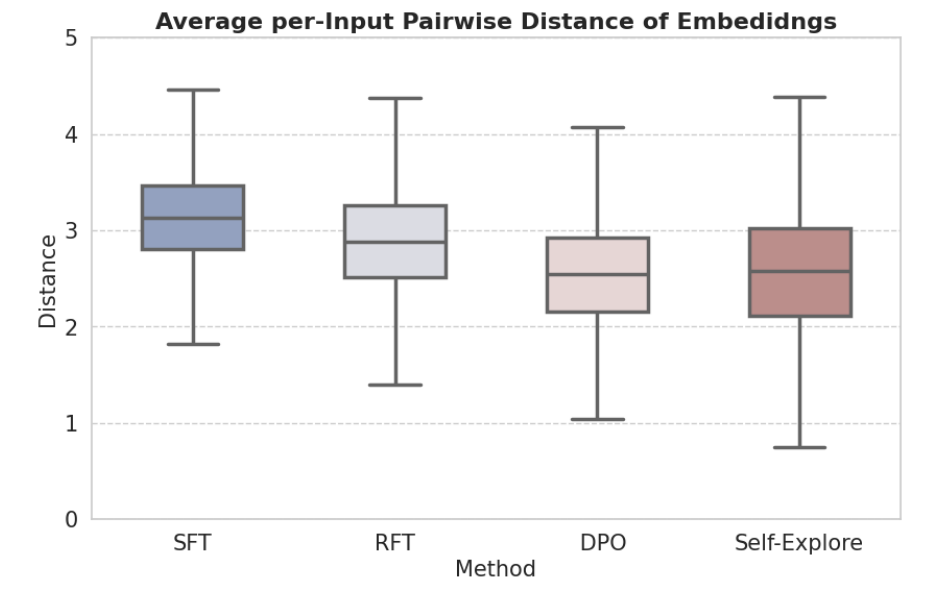}
\caption{Average Per-Input Pairwise Distance of Embeddings of DeepSeek-Math, when trained with different methods.} 
\label{fig:boxplot}
\end{figure}

We plot the distribution of $d_i$ for each training method in the boxplot shown in Figure~\ref{fig:boxplot}. We observe a general decrease in embedding distances from left to right. Particularly DPO and \textsc{Self-Explore} display lower embedding distance than SFT and RFT, hinting at relatively reduced diversity. This phenomena also explains why Pass@K for SFT and RFT is higher compared to those trained with preference learning objective, as SFT and RFT may engage in more \textit{exploration} during test-time. 

In addition, it is important to recognize that a policy characterized by reduced diversity may exhibit limited generalization capabilities, which could be seen as a drawback. Note that for Self-Consistency (maj@k), RFT and SFT surpasses \textsc{Self-Explore} at $K$= 6 and 15, respectively. We find that the reason for this phenomena is due to the concentration of the answer space stemming from the lack of solution diversity, as demonstrated in Appendix \ref{appendix:ans}.

Our models trained with preference learning tend to heavily favor what they identify as an \textit{optimal} answer. Specifically, the reward accuracy when training these models quickly converge to 1, which is illustrated in Appendix \ref{appendix:dpo}, indicating a potential reward exploitation that may lead to limitation in the model's ability to generalize. We hypothesize that this stems from self-training focused on exploring a confined \textit{solution} space, which may not effectively extend to a broader \textit{question} space.

Consequently, the solution distribution becomes skewed, leading to the emergence of overly confident peaks (modes) that may accurately represent the training data but fail to generalize to new unseen questions during testing, as shown by the reduced diversity. In contrast, models trained with SFT or RFT adopt a more \textit{uniform} distribution across potential answers, whereas marginalizing over answers allow for slightly more pronounced peaks to be observable. (\textit{i.e.} Self-Consistency) Overall, these benefits appear diminished when training with preference learning objective with self-generated data. 

In fact, we also observe a similar pattern for the solution distribution in GSM8K. There is also less reduction in total accuracy with increasing $k$ for models trained with preference learning objective. This again can be explained as a \textit{risk} minimization behavior. Regarding the other two metrics, we see that SFT and RFT models exhibit lower performance at lower $k$ values, but they eventually converge to the similar level (maj@k) or even surpass (pass@k) with increasing $k$. We hypothesize this trend again reflects the reduction in diversity within the model's predictions for DPO and \textsc{Self-Explore}.   At the same time, we see that for DPO, rather the performance increases with inclusion of lower-k predictions. This indicates a potential misalignment, which explains the need for the granular supervision during training for a better learning signal.

\section{Answer Distribution}
\label{appendix:ans}

\begin{figure*}[t]
\centering
\includegraphics[width=0.9\textwidth,height=\textheight,keepaspectratio]{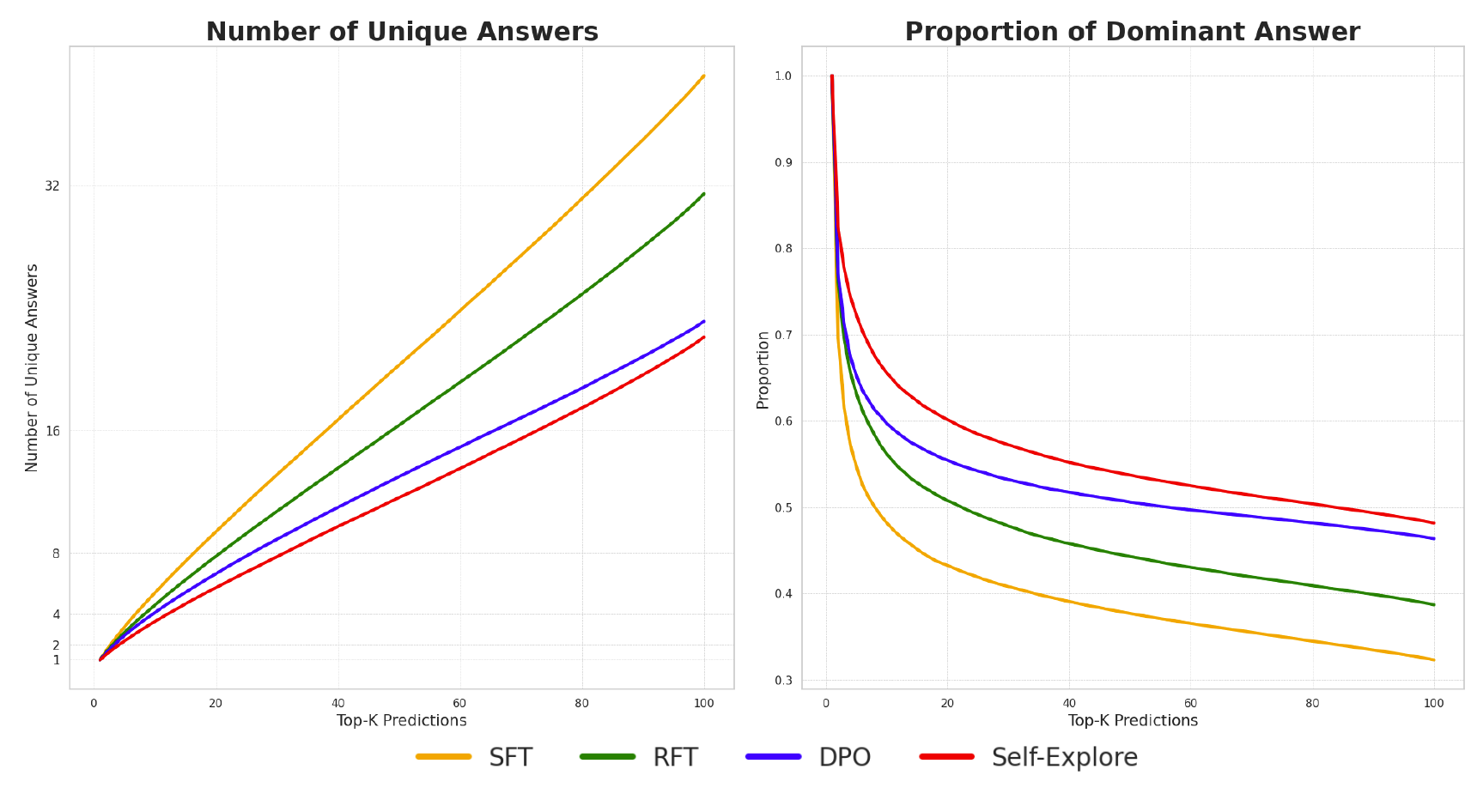}
 \caption{Final answer diversity and the proportion of the dominant (most common) answer within top-k predictions of DeepSeek-Math on MATH Test dataset.}
\label{fig:ans_stats}
\end{figure*}

On the left side of Figure \ref{fig:ans_stats}, we see that the number of unique answers decrease in order of SFT, RFT, DPO, then \textsc{Self-Explore}. Meanwhile on the right, we see that DPO and \textsc{Self-Explore} shows the highest proportion of dominant answer, suggesting a concentrated or skewed distribution of the answer space. These observations support the hypothesis that the model may exhibit overconfidence in its 'optimal' answers, when applied preference learning with self-generated solutions. Such confidence, without sufficient generalization power, could indicate potential overfitting to the training data.

\section{Pairwise Dataset Formation}
\label{appendix:pair}

\subsection{Maximum Pair Constraints}
We initially set no upper limit on the number of response pairs per problem in our dataset. However, preliminary analysis suggested that problems with nearly balanced correct and incorrect responses could potentially generate disproportionately many pairs, risking data overfitting. Thus, we have decided to adopt a maximum threshold of eight pairs (N=8) for each problem $x_i$. While we did not observe such cases many times, we adopted this strategy to ensure a more equitable distribution across different questions.

\subsection{Excluding Conclusion}
When we first ran DPO training, we observed model performance significantly degrading when including \textit{Conclusion} part within the rejected sample (Figure \ref{fig:example_figure}). In such case, our trained model was frequently presenting self-contradictory statements in the conclusion, yielding random answers that were unrelated to the reasoning presented in the preceding steps. We believe it is due to the concurrent presence of definitive statements like "The answer is X" in the chosen and "The answer is Y" in the rejected sample, causing model confusion during training. Therefore we decided to omit the conclusion section (+ \textit{eos} token) from all \textbf{rejected} samples.

\section{Training Dataset Size}
\label{appendix:dataset}
In this section, we discuss about the dataset size utilized for each training method and model. Despite the seemingly comparable amount of training samples (Table \ref{tab:data_size}), we highlight several observations based on the proportion of \textit{question-level} unique instances in each dataset, which is shown in Table \ref{tab:prop1} and \ref{tab:prop2}:

\textbf{1. Few \textit{incorrect} samples for GSM8K} Transitioning from the RFT to the paired dataset, there is a notable reduction in the number of unique questions for GSM8K compared to MATH. This occurred because in several instances, the model generated all 100 solutions correctly, or there were fewer than four incorrect solutions. This overall hints at the scarcity of generated incorrect samples when training with GSM8K dataset.

\textbf{2. Few \textit{correct} samples for MATH}  Despite the model achieving high pass@k rates on the training set (over 90\% for GSM8K and over 70\% for MATH), the actual number of instances that \textit{pass} is notably small for the MATH dataset. Especially, there is a large decline in Table \ref{tab:prop2} when considering the number of unique questions with more than 4 instances. This suggests that for many questions, the models barely reach the correct answer within 100 generations.

\section{DPO Training}
\label{appendix:dpo}
\begin{figure*}[t]
\centering
\includegraphics[width=1\textwidth,height=\textheight,keepaspectratio]{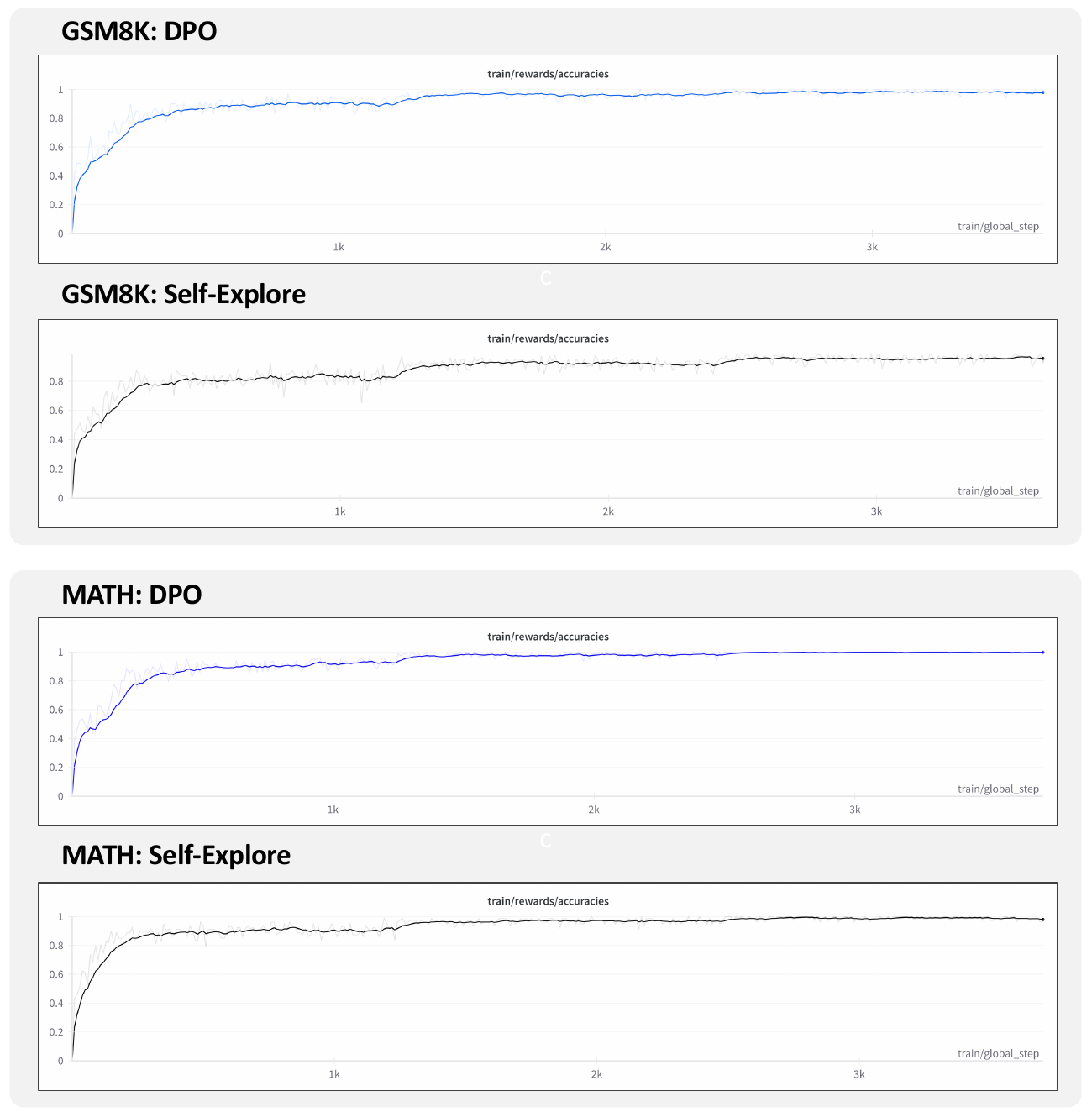}
 \caption{Reward accuracy (\textit{i.e.} winrate of \textit{chosen} over \textit{rejected} samples) of DPO and \textsc{Self-Explore} during training of DeepSeek-Math. For both methods, the accuracy quickly converges to 1 regardless of the supervision type.}
\label{fig:dpo}
\end{figure*}

While in the original paper~\citep{rafailov2023direct}, DPO training displayed chosen completion's win rate over the rejected completion around $60\text{-}70\%$, we observe in Figure \ref{fig:dpo} that the reward of \textit{chosen} sample quickly suprasses that of \textit{rejected} in our early stages, with winrate converging to 1 in both datasets. We hypothesize that this occurs for two reasons. 1) Chosen completion is generated by $M_\mathrm{RFT}$ which is closer to the target model's distribution, while rejected is generated by $M_\mathrm{SFT}$. 2) Models can also quickly learn to distinguish the preference within the limited question numbers, which may nonetheless lead to overfitting.

\section{Examples}

In Figure \ref{fig:flask_sample}, we see that while the DPO model concludes prematurely after the 5th step, falling into a \textit{pit}, the \textsc{Self-Explore} model continues to generate subsequent steps robustly, ultimately arriving at the correct answer. This sample effectively illustrates how our method achieves step-level robustness through targeted step-level supervision.

\section{FLASK Prompt}
\label{appendix:flask}

\begin{figure*}[t]
\centering
\includegraphics[width=1\textwidth,height=\textheight,keepaspectratio]{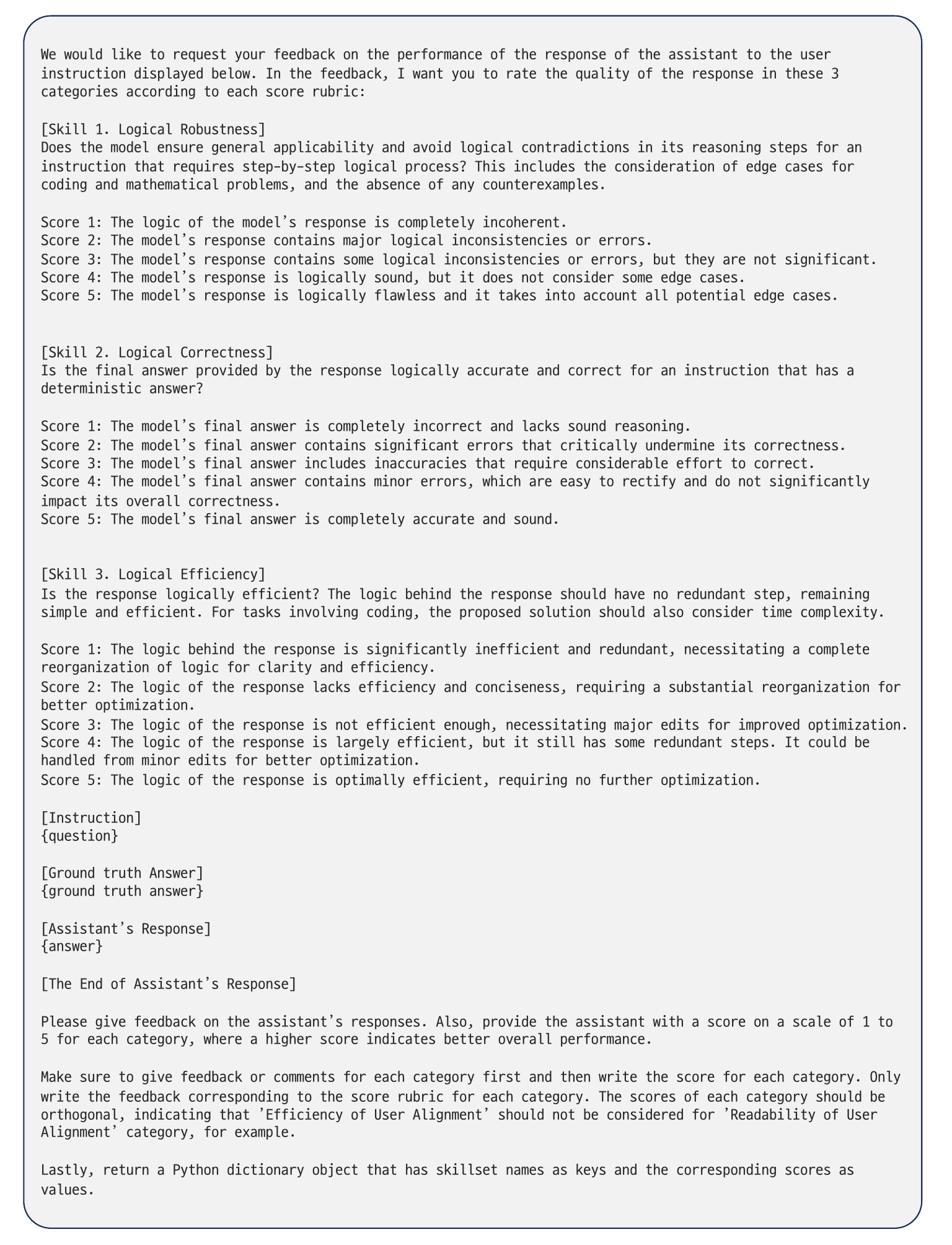}
 \caption{GPT-4 prompt used for FLASK evaluation.}
\label{fig:flask_prompt}
\end{figure*}

In Figure~\ref{fig:flask_prompt}, we present the prompt used for the GPT-4 FLASK evaluation, which assesses three key logical skills: robustness, correctness, and efficiency. These skills are evaluated against the ground truth (GT) solution using a deterministic rubric for each criterion. 

\begin{figure*}[t]
\centering
\includegraphics[width=1\textwidth,height=\textheight,keepaspectratio]{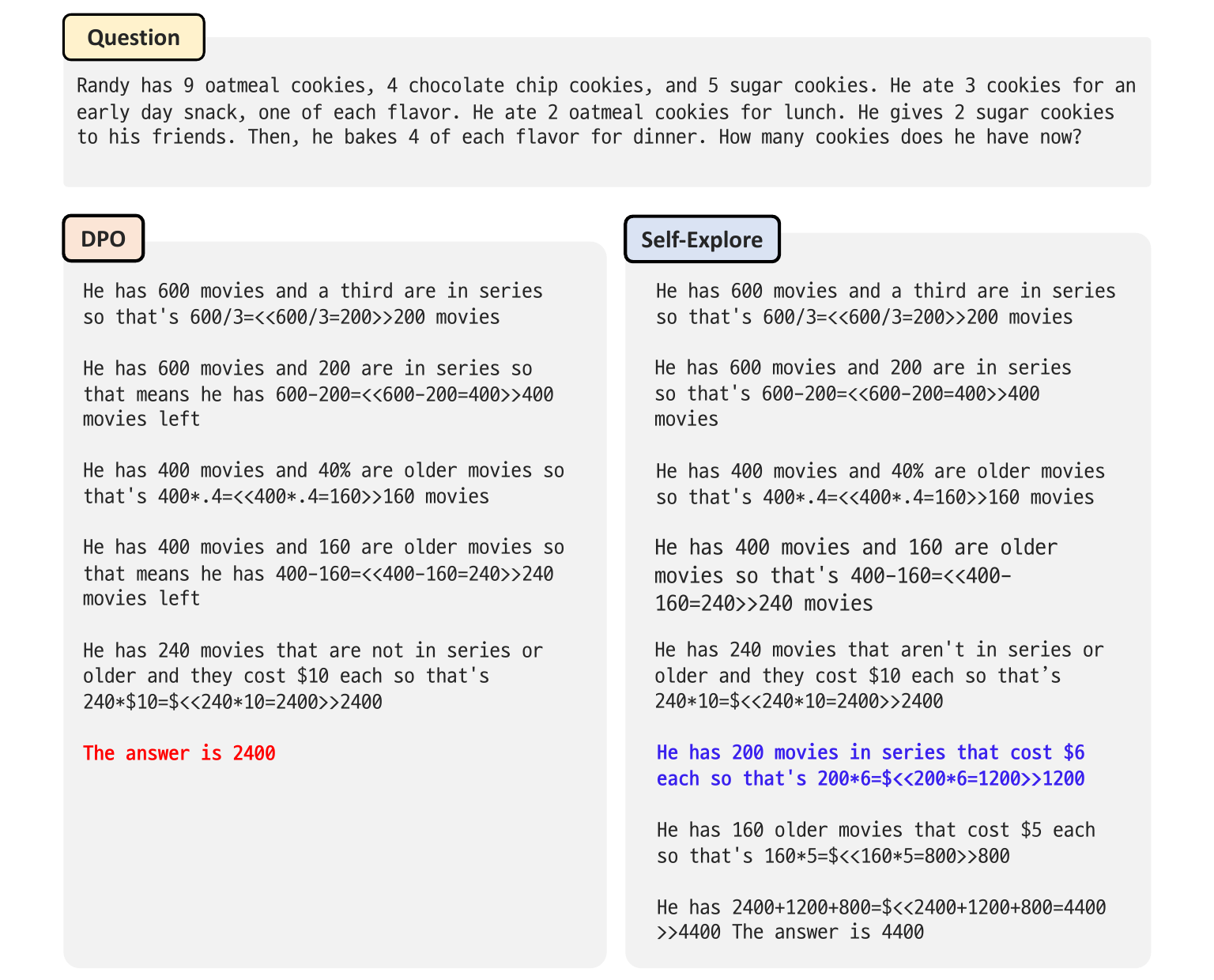}
 \caption{Examples for sample generated by DPO and Self-Explore, respectively.}
\label{fig:flask_sample}
\end{figure*}

\begin{figure*}[t]
\centering
\includegraphics[width=1\textwidth,height=\textheight,keepaspectratio]{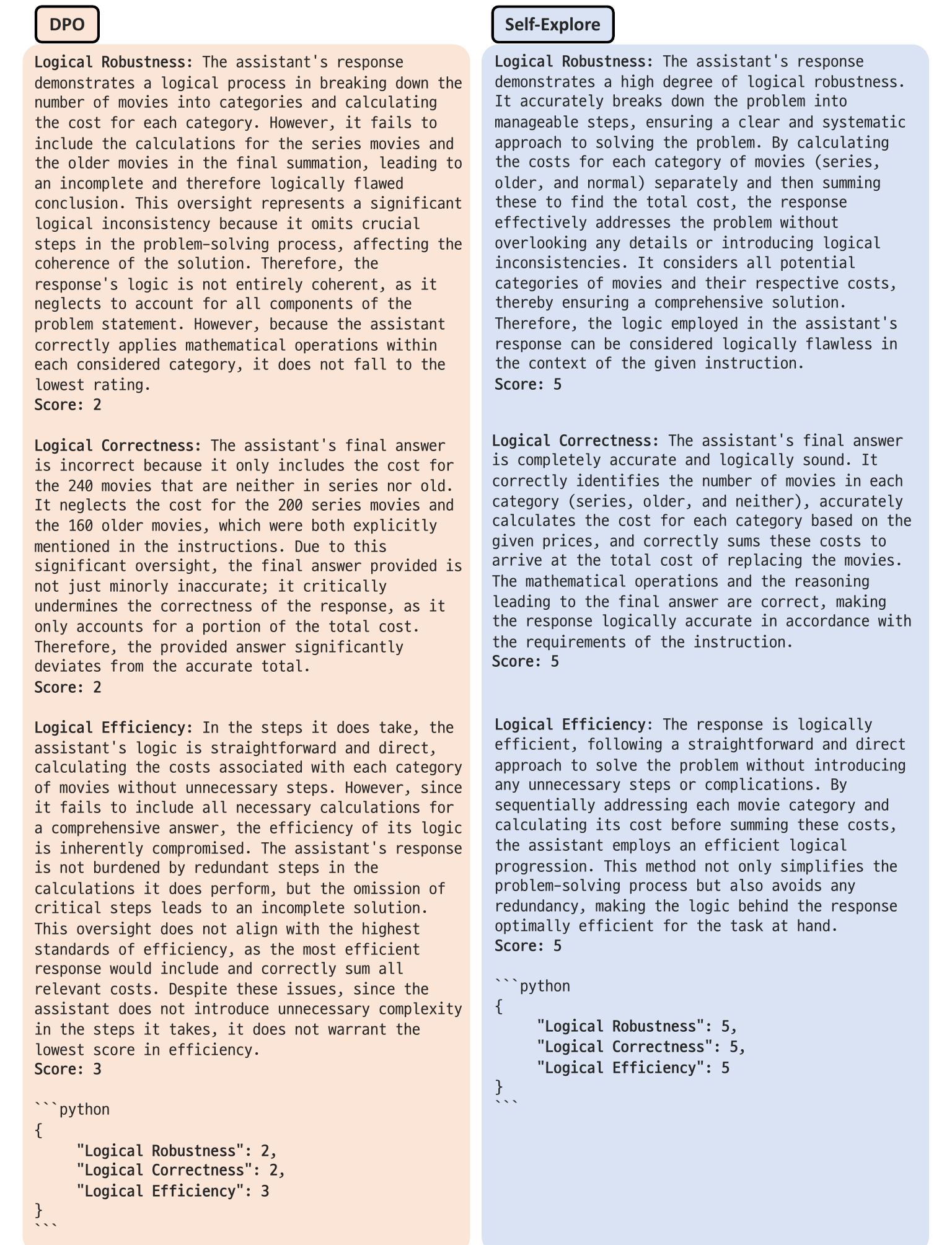}
 \caption{Results of GPT-4 FLASK evaluation for the generated solutions shown in Figure~\ref{fig:flask_sample}.}
\label{fig:flask_eval}
\end{figure*}

When evaluating the example responses present in Figure~\ref{fig:flask_sample}, we see that DPO model receives a score of $2, 3, 2$ while \textsc{Self-Explore} gets a full score of $5, 5, 5$ for Logical robustness, Correctness and Efficiency, respectively, as shown in Figure~\ref{fig:flask_eval}. GPT-4's coherent explanation of these scores adds credibility to the overall FLASK evaluation result in Table~\ref{tab:logical}, underscoring the superior quality of responses generated by the \textsc{Self-Explore} model.

\section{Computational Costs}
\label{appendix:compute}
We report the overall computational costs (\textsc{i.e.} GPU Hours) for each baseline, including the exploration stage in Table \ref{tab:gpu_hours}. Our baselines involve different training (Tr.) and generation (Gen.) stages. Note that the table reflects the following configuration:

\begin{itemize}
    \item Mistral 7B
    \item 5 epoch SFT \& RFT training
    \item 3 epoch DPO training
    \item 7.4M RFT samples generated
    \item 39K pairwise-data exploration
\end{itemize}

\renewcommand{\arraystretch}{1.1}
\begin{table*}
\small
\centering
\begin{tabularx}{0.9\textwidth}{@{}X >{\raggedleft\arraybackslash}X >{\raggedleft\arraybackslash}X >{\raggedleft\arraybackslash}X@{}}
\toprule
\textbf{\; Methods} & \textbf{Mistral}   & \textbf{Llemma}   & \textbf{DeepSeek} \\ \midrule \midrule
\multicolumn{4}{l}{\textbf{Dataset: GSM8K}}                          \\  \midrule   
\; FT     & 7,473 & 7,473 & 7,473                \\
\; RFT    & 67,755*            & 38,989            & 52,005            \\
\; pair   & 56,443*            & 37,058            & 38,872            \\
\; g-pair & 56,283*            & 36,812            & 38,618            \\ \midrule
\multicolumn{4}{l}{\textbf{Dataset: MATH}}                                \\ \midrule
\; FT     & 7,500 & 7,500 & 7,500                               \\
\; RFT    & 31,839            & 34,419            & 40,654            \\
\; pair   & 31,527            & 34,124            & 39,769            \\
\; g-pair & 31,248            & 33,960            & 39,496            \\ \bottomrule
\end{tabularx}
\caption{Dataset size used for each training method, by each model.}
\label{tab:data_size}
\text{* denotes no maximum pair formation constraint}
\end{table*}

\renewcommand{\arraystretch}{1.1}
\begin{table*}
\small
\centering
\begin{tabularx}{0.9\textwidth}{@{}X >{\raggedleft\arraybackslash}X >{\raggedleft\arraybackslash}X >{\raggedleft\arraybackslash}X@{}}
\toprule
& \textbf{Mistral}   & \textbf{Llemma}   & \textbf{DeepSeek} \\ \midrule
\hspace{1em}FT     & 1.0    & 1.0  & 1.0    \\ \midrule
\multicolumn{4}{l}{\textbf{\textsc{Number of Samples $\geq 1$}}}                             \\ \midrule
\hspace{1em}RFT    & 0.9830    & 0.9252    & 0.9917   \\
\hspace{1em}pair   & 0.9213    & 0.9113    & 0.8955   \\
\hspace{1em}g-pair & 0.9212    & 0.9098    & 0.8947   \\ \midrule
\multicolumn{4}{l}{\textbf{\textsc{Number of Samples $\geq 4$}}}                               \\ \midrule
\hspace{1em}RFT    & 0.7376    & 0.7281    & 0.8616   \\
\hspace{1em}pair    & 0.6204    & 0.5739    & 0.6063   \\
\hspace{1em}g-pair & 0.6195    & 0.5700    & 0.6024   \\ \bottomrule
\end{tabularx}
\caption{Proportion of questions in GSM8K with at least N instances for each training method, by each model.}
\label{tab:prop1}
\end{table*}

\renewcommand{\arraystretch}{1.1}
\begin{table*}
\small
\centering
\begin{tabularx}{0.9\textwidth}{@{}X >{\raggedleft\arraybackslash}X >{\raggedleft\arraybackslash}X >{\raggedleft\arraybackslash}X@{}}
\toprule
       & \textbf{Mistral}   & \textbf{Llemma}   & \textbf{DeepSeek} \\ \midrule
\hspace{1em}FT     & 1.0    & 1.0  & 1.0                  \\ \midrule
\multicolumn{4}{l}{\textbf{\textsc{Number of Samples $\geq 1$}}}                             \\ \midrule
\hspace{1em}RFT    & 0.7345    & 0.7587    & 0.8240   \\
\hspace{1em}pair   & 0.7345    & 0.7356    & 0.8225   \\
\hspace{1em}g-pair & 0.7345    & 0.7353    & 0.7971   \\ \midrule
\multicolumn{4}{l}{\textbf{\textsc{Number of Samples $\geq 4$}}}                               \\ \midrule
\hspace{1em}RFT    & 0.4904    & 0.5375    & 0.6479   \\
\hspace{1em}pair   & 0.4844    & 0.5320    & 0.6309   \\
\hspace{1em}g-pair  & 0.4819    & 0.5309    & 0.6292   \\ \bottomrule
\end{tabularx}
\caption{Proportion of questions in MATH with at least N instances for each training method, by each model.}
\label{tab:prop2}
\end{table*}

\newpage

\renewcommand{\arraystretch}{1.1} 
\begin{table*}
\small
\centering
\begin{tabularx}{0.9\textwidth}{@{}X >{\raggedleft\arraybackslash}X >{\raggedleft\arraybackslash}X >{\raggedleft\arraybackslash}X >{\raggedleft\arraybackslash}X >{\raggedleft\arraybackslash}X >{\raggedleft\arraybackslash}X@{}}
\toprule
\textbf{Stages}         & \textbf{SFT (Tr.)} & \textbf{RFT (Gen.)} & \textbf{RFT (Tr.)} & \textbf{Exploration (Gen.)} & \textbf{DPO (Tr.)} & \textbf{Total Time (hr)} \\ \midrule
\textbf{GPU Hours} & 1.3                & 6                  & 11.2               & 2.7                         & 20                 &                          \\ \midrule
\textbf{SFT}            & \checkmark         &                    &                    &                             &                    & 1.3                      \\
\textbf{RFT}            & \checkmark         & \checkmark         & \checkmark         &                             &                    & 18.5                     \\
\textbf{DPO}            & \checkmark         & \checkmark         & \checkmark         &                             & \checkmark         & 38.5                     \\
\textbf{Self-Explore}   & \checkmark         & \checkmark         & \checkmark         & \checkmark                  & \checkmark         & 41.2                     \\ \bottomrule
\end{tabularx}
\caption{GPU hours for different baselines.}
\label{tab:gpu_hours}
\end{table*}

\end{document}